\documentclass{article}

\usepackage{PRIMEarxiv}

\usepackage[utf8]{inputenc} 
\usepackage[T1]{fontenc}    
\usepackage{hyperref}       
\usepackage{url}            
\usepackage{booktabs}       
\usepackage{amsfonts}       
\usepackage{nicefrac}       
\usepackage{microtype}      
\usepackage{lipsum}
\usepackage{fancyhdr}       
\usepackage{graphicx}       
\usepackage{colortbl,xcolor}
\usepackage{bm}
\usepackage{amsmath}
\graphicspath{{media/}}     

\title{GroupKAN: Efficient Kolmogorov-Arnold Networks via Grouped Spline Modeling
}

\author{
  Guojie Li\thanks{These authors contributed equally to this work.} \\
  School of Robotics, Xi'an Jiaotong-Liverpool University \\
  Suzhou, Jiangsu 215000, China \\
  Department of Computer Science, University of Liverpool, Liverpool L69 3BX, UK \\
  \And
  Tianyi Liu\footnotemark[1] \\
  School of Robotics, Xi'an Jiaotong-Liverpool University \\
  Suzhou, Jiangsu 215000, China \\
  Department of Computer Science, University of Liverpool, Liverpool L69 3BX, UK \\
  \And
  Anwar P.P.\ Abdul Majeed \\
  School of Robotics, Xi'an Jiaotong-Liverpool University, Suzhou, Jiangsu 215000, China \\
  Faculty of Engineering and Technology, Sunway University, Selangor Darul Ehsan 47500, Malaysia \\
  \And
  Muhammad Ateeq \\
  School of Internet of Things, Xi'an Jiaotong-Liverpool University \\
  Suzhou, Jiangsu 215000, China \\
  \And
  Anh Nguyen \\
  Department of Computer Science, University of Liverpool \\
  Liverpool L69 3BX, United Kingdom \\
  \And
  Fan Zhang\thanks{Corresponding author: \href{mailto:Fan.Zhang@xjtlu.edu.cn}{Fan.Zhang@xjtlu.edu.cn}} \\
  School of Robotics, Xi'an Jiaotong-Liverpool University \\
  Suzhou, Jiangsu 215000, China
}

\begin{document}
\maketitle

\begin{abstract}
    
Medical image segmentation demands models that achieve high accuracy while maintaining computational efficiency and clinical interpretability. 
While recent Kolmogorov-Arnold Networks (KANs) offer powerful adaptive non-linearities, their full-channel spline transformations incur a quadratic parameter growth of $\mathcal{O}(C^{2}(G+k))$ with respect to the channel dimension $C$, where $G$ and $k$ denote the number of grid intervals and spline polynomial order, respectively. 
Moreover, unconstrained spline mappings lack structural constraints, leading to excessive functional freedom, which may cause overfitting under limited medical annotations.
To address these challenges, we propose GroupKAN (Grouped Kolmogorov-Arnold Networks), an efficient architecture driven by group-structured spline modeling. 
Specifically, we introduce: (1) Grouped KAN Transform (GKT), which restricts spline interactions to intra-group channel mappings across $g$ groups, effectively reducing the spline-induced quadratic expansion to \textbf{$\mathcal{O}(C^2(\frac{G+k}{g} + 1))$}, thereby significantly lowering the effective quadratic coefficient; 
and (2) Grouped KAN Activation (GKA), which applies shared spline functions within each group to enable efficient token-wise non-linearities. 
By imposing structured constraints on channel interactions, GroupKAN achieves a substantial reduction in parameter redundancy without sacrificing expressive capacity.
Extensive evaluations on three medical benchmarks (BUSI, GlaS, and CVC) demonstrate that GroupKAN achieves an average IoU of 79.80\%, outperforming the strong U-KAN baseline by +1.11\% while requiring only 47.6\% of the parameters (3.02M vs. 6.35M). Qualitative results further reveal that GroupKAN produces sharply localized activation maps that better align with the ground truth than MLPs and KANs, significantly enhancing clinical interpretability. \textbf{Code available at:} \url{https://github.com/liguojie09/GroupKAN}
\end{abstract}

\section{Introduction}
\label{sec:intro}



Medical image segmentation is critical for clinical diagnosis and treatment planning~\cite{ramesh2021review,wang2022medical}. Historically, the field has been dominated by Convolutional Neural Networks (CNNs) \cite{ronneberger2015u,zhou2018unet++} and Vision Transformers (ViTs) \cite{gu2023mamba}. However, these methods often force a trade-off between limited local receptive fields and the heavy computational burdens required for global context modeling \cite{fournier2023practical,hua2022transformer,shuvo2022efficient,younesi2024comprehensive}. Recently, Kolmogorov-Arnold Networks (KANs)~\cite{liu2024kan} have emerged as an alternative, replacing static linear weights with learnable univariate splines to capture complex non-linearities dynamically~\cite{bodner2024convolutional,drokin2024kolmogorov}. Building on this paradigm shift, architectures like U-KAN~\cite{li2025u} have demonstrated impressive representational capabilities by integrating KAN layers into a standard U-shaped backbone.

However, the direct application of unconstrained KANs in medical imaging introduces computational and theoretical bottlenecks. 
Specifically, U-KAN’s full-channel spline transformations lead to a quadratic parameter growth of $\mathcal{O}(C^{2}(G + k))$, where $C$, $G$, and $k$ denote the channel dimension, grid intervals, and spline order, respectively.
This quadratic coefficient expansion substantially increases parameter redundancy, restricting
its deployment in resource-constrained clinical environments.
On the theoretical side, the unconstrained nature of these spline mappings lacks structural limitations, creating an excessively flexible hypothesis space. 
In statistical learning theory, a high degree of freedom corresponds to a large Vapnik-Chervonenkis (VC) dimension $h$, which reflects a model's capacity to memorize complex patterns \cite{li2025generalization,vapnik1999overview}. 
While high capacity is beneficial for large-scale datasets, it can become problematic in medical segmentation tasks, where annotated data is notoriously scarce \cite{tajbakhsh2020embracing}. 
Driven by this inflated VC dimension $h$, such models face the risk of overfitting. 
Therefore, a structural constraint is urgently needed to regularize the functional hypothesis space, effectively lowering $h$ without sacrificing the adaptive power of KANs.

To address these limitations, we propose GroupKAN, a lightweight architecture beyond U-KAN, driven by group-structured spline modeling. Inspired by the parallel processing pathways of retinal ganglion cells, we demonstrate that channel grouping serves as a highly effective structural constraint for unconstrained KANs \cite{wassle2004parallel}. Specifically, we introduce two novel components: (1) Grouped KAN Transform (GKT), which partitions channels into $g$ independent groups to model intra-group channel dependencies via multivariate spline mappings. 
This structural design reduces the spline-induced quadratic expansion from $\mathcal{O}(C^2(G+k))$ to $\mathcal{O}(C^2(\frac{G+k}{g} + 1))$, thereby significantly lowering the effective quadratic coefficient while preserving full channel connectivity.
(2) Grouped KAN Activation (GKA), where all channels within the same group share identical spline functions. This shared mechanism not only enables efficient token-wise non-linearities but also regularizes the parameter space. Together, GKT and GKA effectively compress the VC dimension $h$, forcing the model to discard spurious noise.
Extensive evaluations on BUSI, GlaS, and CVC-ClinicDB benchmarks validate the superiority of GroupKAN. It achieves an average IoU of 79.80\%, outperforming U-KAN by +1.11\% while operating with merely 47.6\% of its parameters. Furthermore, qualitative analyses confirm that our structural priors force the model to yield localized activation maps that accurately align with the ground truth.

Our contributions are summarized as follows:
(1) We propose GroupKAN, a parameter-efficient Kolmogorov-Arnold backbone for medical image segmentation, composed of Grouped KAN Transform (GKT) and Grouped KAN Activation (GKA) to enable group-wise spline modeling.
(2) By introducing channel grouping into spline mappings, GroupKAN reduces the effective quadratic coefficient of full-channel transformations, lowering parameter redundancy and improving generalization under limited supervision.
(3) Extensive evaluations on three benchmarks demonstrate that GroupKAN achieves superior accuracy-efficiency trade-offs compared with existing backbones.


\begin{figure}[t]
\centering
\includegraphics[width=\textwidth]{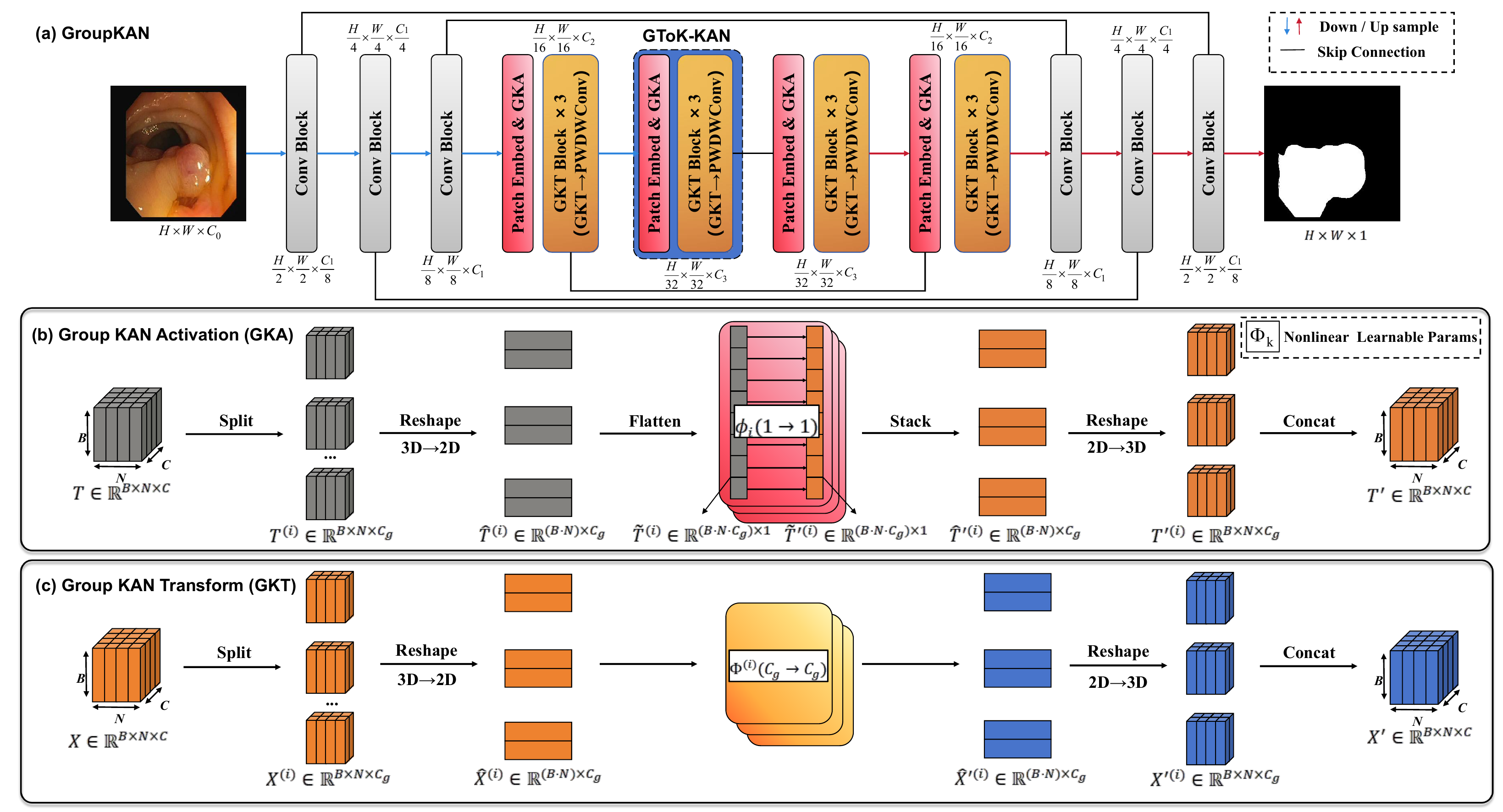} 
\caption{Overview of GroupKAN. The network follows an encoder-decoder design with Group ToK-KAN blocks integrated at the bottleneck for feature modeling.}
\label{method}
\end{figure}

\section{Method}
\subsection{Preliminaries}
\label{sec:preliminaries}
To enhance non-linear capacity, Kolmogorov-Arnold Networks (KANs) \cite{liu2024kan} replace static linear weights with learnable 1D B-spline functions. Specifically, each output channel $y_q$ is computed by aggregating transformed features from all input channels $x_p$ via distinct spline mappings $\phi_{q,p}$, as formulated in Eq.~\ref{eq:kan}.
\begin{equation}
\label{eq:kan}
y_{q}=\sum_{p=1}^{C}\phi_{q,p}(x_{p})
\end{equation}

\subsection{Overall Architecture of GroupKAN}
As illustrated in Fig.~\ref{method}(a), GroupKAN adopts a highly structured U-shaped architecture tailored for dense medical image segmentation. A three-stage convolutional encoder progressively downsamples the spatial resolution while extracting hierarchical features with increasing channel capacities of $C_1/8$, $C_1/4$, and $C_1$. At the bottleneck, two stacked Group Tokenized KAN (GToK-KAN) blocks operate at intermediate dimensions $C_2$ and $C_3$ to efficiently capture global context. These core blocks tokenize the convolutional features and employ learnable spline-based operators to perform structured token-wise activations and channel-wise transformations. Finally, a symmetric convolutional decoder progressively upsamples the feature maps, integrating them with the corresponding encoder representations via skip connections across dimensions $C_3$ through $C_1/8$, before a $1 \times 1$ convolution predicts the final segmentation mask.\\
\\
\noindent\textbf{Group Tokenized KAN (GToK-KAN) Blocks.}
Each GToK-KAN block transforms encoder features into tokenized representations and applies group-aware functional modules, explicitly decoupled into token-wise activation (GKA) and intra-group channel transformation (GKT), as detailed in Fig.~\ref{method}(b) and (c).\\
\\
\noindent\textbf{Group Tokenized Representation.}
Let $X \in \mathbb{R}^{B \times C_0 \times H \times W}$ denote the input feature map from the encoder, we first convert it into a sequence of tokens suitable for functional modeling. A $3 \times 3$ convolution with stride 2 is applied to $X$, producing an intermediate feature map $F \in \mathbb{R}^{B \times C \times H' \times W'}$. The spatial dimensions are then flattened to form a token sequence $T \in \mathbb{R}^{B \times N \times C}$, where $N = H' \cdot W'$. Each token corresponds to a local receptive field, retaining spatial context through convolutional embedding. This tokenized representation provides a unified interface for subsequent group-wise functional modules. \\
\\
\noindent\textbf{Grouped KAN Activation (GKA).}
Standard point-wise activations~\cite{wu2021cvt,wang2022pvt} may ignore channel structures. Our GKA introduces efficient, group-aware nonlinearities. The input $T \in \mathbb{R}^{B \times N \times C}$ is partitioned into $g$ groups 
of size $C_g = C/g$, yielding $T^{(i)} \in \mathbb{R}^{B \times N \times C_g}$ ($i=1, \dots, g$). To enable intra-group sharing, each group is first reshaped into a 2D matrix $\hat{T}^{(i)} \in \mathbb{R}^{(B \cdot N) \times C_g}$ and then flattened into a 1D vector $\tilde{T}^{(i)} \in \mathbb{R}^{(B \cdot N \cdot C_g) \times 1}$. Subsequently, a single shared 1D spline $\phi_i$ is applied element-wise to this flattened vector, as formulated in Eq.~\ref{eq:gka}, ensuring that all channels within the same group share identical spline functions:
\begin{equation}
\label{eq:gka}
\tilde{T}^{\prime(i)} = \phi_i(\tilde{T}^{(i)})
\end{equation}
The activated outputs are sequentially reshaped back to an intermediate 2D matrix $\hat{T}^{\prime(i)} \in \mathbb{R}^{(B \cdot N) \times C_g}$ and then to the 3D tensor $T^{\prime(i)} \in \mathbb{R}^{B \times N \times C_g}$. Finally, they are concatenated across all groups to form the activated sequence $T^{\prime} \in \mathbb{R}^{B \times N \times C}$. This intra-group sharing severely bounds the parameter space while preserving nonlinear capacity without introducing additional computational overhead.
\\
\\
\noindent\textbf{Grouped KAN Transform (GKT).}
\label{sec:gkt}
To model intra-group dependencies, the input sequence $X \in \mathbb{R}^{B \times N \times C}$ (e.g., from GKA) is partitioned into $g$ groups denoted as $X^{(i)} \in \mathbb{R}^{B \times N \times C_g}$ ($i=1, \dots, g$). Each group is reshaped into a 2D matrix $\hat{X}^{(i)} \in \mathbb{R}^{(B \cdot N) \times C_g}$ and passed through a learnable nonlinear mapping. Specifically, a dense $C_g \rightarrow C_g$ multivariate spline mapping $\Phi^{(i)}$ is applied:
\begin{equation}
\label{eq:gkt}
\hat{X}^{\prime(i)} = \Phi^{(i)}(\hat{X}^{(i)}), \quad \Phi^{(i)} = \left[\phi_1^{(i)}, \dots, \phi_{C_g}^{(i)}\right]
\end{equation}
The transformed outputs $\hat{X}^{\prime(i)} \in \mathbb{R}^{(B \cdot N) \times C_g}$ are reshaped back to 3D tensors $X^{\prime(i)} \in \mathbb{R}^{B \times N \times C_g}$ and concatenated to form the complete representation $X^{\prime} \in \mathbb{R}^{B \times N \times C}$. To enable cross-group routing and spatial refinement, $X^{\prime}$ then passes through $1 \times 1$ pointwise and $3 \times 3$ depthwise convolutions. 

\subsection{Computation Cost}
\label{sec:computation_cost}
In standard architectures, cross-channel mixing via Multi-Layer Perceptrons (MLPs) requires $\mathcal{O}(C^2)$ parameters for $C$ channels, as it relies on a dense transformation matrix ${W} \in \mathbb{R}^{C \times C}$ to fully connect all input and output channels. While the dense pairwise mapping of standard KANs significantly boosts representational power, it causes the overall parameter count to balloon to $\mathcal{O}(C^2(G + k))$. Here, $G$ represents the number of grid intervals partitioning the input domain, and $k$ denotes the piecewise polynomial order of the spline. In B-spline theory \cite{liu2024kan}, parameterizing a continuous curve over $G$ spatial intervals strictly requires a linear combination of exactly $G+k$ basis functions to maintain local polynomial flexibility of order $k$. Since each basis function must be weighted by a distinct learnable coefficient, this dense parameterization becomes computationally prohibitive in deep medical networks where $C$ is inherently large. 

To break this bottleneck, our proposed Grouped KAN Transform (GKT, detailed in Sec.~\ref{sec:gkt}) partitions the $C$ channels into $g$ groups 
and performs intra-group spline mapping, reducing the spline-induced parameter complexity from $\mathcal{O}(C^2(G+k))$ to $\mathcal{O}(C^2(\frac{G+k}{g}))$.
Subsequently, a $1 \times 1$ pointwise convolution is applied to restore inter-group communication, introducing an additional $\mathcal{O}(C^2)$ parameter cost.
This structural constraint substantially reduces the spline-induced quadratic expansion while maintaining efficient cross-group mixing, yielding the refined complexity formulated as $\mathcal{O}\left(C^{2}\left(\frac{G+k}{g} + 1\right)\right)$. 
Consequently, in typical settings ($G=5$, $k=3$, $g=16$), the relative parameter complexity follows approximately MLP : GKT : Standard KAN $\approx\ 1 : 1.5 : 8$.

\subsection{Why GroupKAN Works}
The superiority of GToK-KAN stems from resolving the capacity-regularization dilemma. According to Statistical Learning Theory \cite{vapnik1999overview}, the risk $R$ of a learning machine $f$ is upper-bounded by its empircal risk $\hat{R}(f)$ plus a complexity term that scales with the $h$ and sample size $n$: $R(f) \le \hat{R}(f) + \mathcal{O}(\sqrt{h \log(n)/n})$. 
Even if the empirical risk is minimised, the generalization error may remain large when the complexity term is substantial due to a large VC dimension $h$.
Under limited sample size $n$, unconstrained KANs (e.g., U-KAN) with excessively flexible spline parameterizations may exhibit enlarged effective capacity, increasing the risk of overfitting to stochastic imaging noise.
GroupKAN elegantly breaks this bottleneck via the Grouped KAN Transform (GKT). By partitioning channels into $g$ groups, GKT imposes a block-diagonal constraint that restricts highly flexible splines to intra-group mappings,  
effectively constraining the hypothesis space and reducing the non-linear VC dimension ($h_{GroupKAN} \ll h_{UKAN}$), i.e., reducing the effective model capacity relative to unconstrained KANs.

\section{Experiments and Results}
\subsection{Datasets and Implementation Details}
Following the exact protocol of U-KAN~\cite{li2025u} for fair comparison, we evaluate GroupKAN on three segmentation datasets: BUSI~\cite{al2020dataset}, GlaS~\cite{sirinukunwattana2017gland}, and CVC~\cite{bernal2015wm}.
The networks were trained for 400 epochs (batch size 8) with standard spatial augmentations. We optimized the network using Adam with the BCE-Dice loss \cite{milletari2016v}, applying a cosine-annealed learning rate decaying from $1\times10^{-4}$ to $1\times10^{-5}$. Crucially, to ensure rigorous evaluation, our main results (Table~\ref{tab:seg_comparison}) are averaged across three independent runs with varying random seeds and 80\%/20\% data splits.
FLOPs are consistently measured at a $512\times512$ resolution. 
\begin{table*}[t]
\centering
\footnotesize
\setlength{\tabcolsep}{1pt}
\caption{Quantitative segmentation results on BUSI, GlaS, and CVC-ClinicDB among different Method (M): 1:U-Net~\cite{ronneberger2015u}, 2:Att-Unet~\cite{oktay2018attention}, 3:U-Net++~\cite{zhou2018unet++}, 4: U-NeXt~\cite{valanarasu2022unext}, 5:Rolling-UNet~\cite{liu2024rolling}, 6: U-Mamba~\cite{ma2024u}, 7: U-KAN~\cite{li2025u} }
\begin{tabular}{l|cc|cc|cc}
\toprule
\textbf{M} & \multicolumn{2}{c|}{\textbf{BUSI \cite{al2020dataset}}} & 
\multicolumn{2}{c|}{\textbf{GlaS \cite{sirinukunwattana2017gland}}} & 
\multicolumn{2}{c}{\textbf{CVC \cite{bernal2015wm}}} \\
\cmidrule(lr){2-3} \cmidrule(lr){4-5} \cmidrule(lr){6-7}
& IoU↑ & F1↑ & IoU↑ & F1↑ & IoU↑ & F1↑ \\
\midrule
1 & 57.36$\pm$4.61 & 72.41$\pm$3.37 & 86.51$\pm$1.13 & 92.68$\pm$0.82 & 83.66$\pm$1.03 & 91.02$\pm$0.87 \\
2  & 57.62$\pm$3.93 & 72.77$\pm$3.14 & 86.67$\pm$1.25 & 92.77$\pm$0.77 & 84.39$\pm$0.68 & 91.52$\pm$0.34 \\
3 & 58.25$\pm$3.58 & 73.23$\pm$2.76 & 87.13$\pm$0.82 & 93.02$\pm$0.47 & 84.57$\pm$1.33 & 91.66$\pm$0.97 \\
4 & 58.27$\pm$1.56 & 73.38$\pm$1.77 & 83.64$\pm$1.02 & 90.98$\pm$0.52 & 75.62$\pm$1.08 & 86.21$\pm$0.87 \\
5 & 60.23$\pm$1.23 & 75.05$\pm$1.65 & 86.53$\pm$0.87 & 92.50$\pm$0.74 & 83.12$\pm$1.38 & 90.79$\pm$1.12 \\
6 & 61.79$\pm$3.54 & 76.13$\pm$3.27 & 87.08$\pm$0.52 & 93.11$\pm$0.37 & 84.88$\pm$0.63 & 91.74$\pm$0.44 \\
7 & 63.41$\pm$2.67 & 76.51$\pm$2.88 & 87.51$\pm$0.39 & 93.29$\pm$0.18 & 85.14$\pm$0.67 & 91.99$\pm$0.36 \\
\rowcolor{gray!20}
\textbf{Ours} & \textbf{65.18$\pm$2.73} & \textbf{78.22$\pm$2.72} & \textbf{87.91$\pm$0.36} & \textbf{93.66$\pm$0.19} & \textbf{86.31$\pm$0.73} & \textbf{92.35$\pm$0.64} \\
\bottomrule
\end{tabular}
\label{tab:seg_comparison}
\end{table*}

\subsection{Comparison Results}
\begin{figure*}[t]
\centering
\includegraphics[width=1\textwidth]{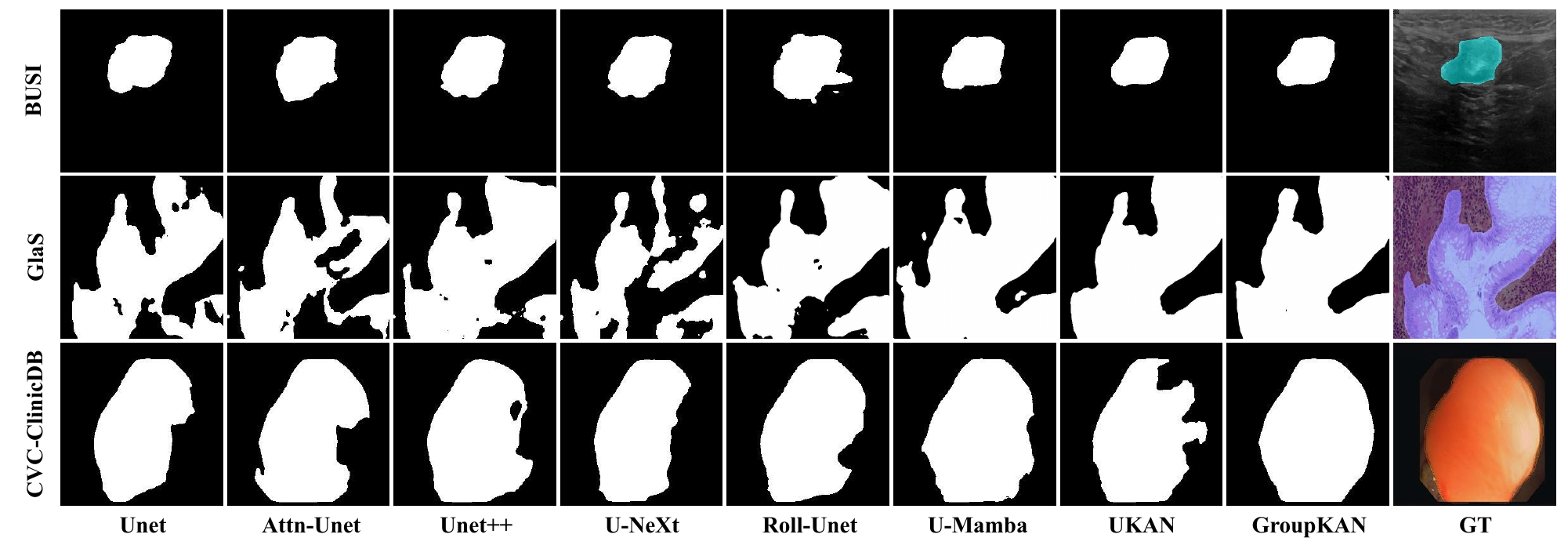} 
\caption{Qualitative comparison of segmentation results.}
\label{fig:quality}
\end{figure*}

\begin{figure}[t]
\centering
\begin{minipage}[c]{0.48\textwidth}
\centering
\includegraphics[width=\linewidth]{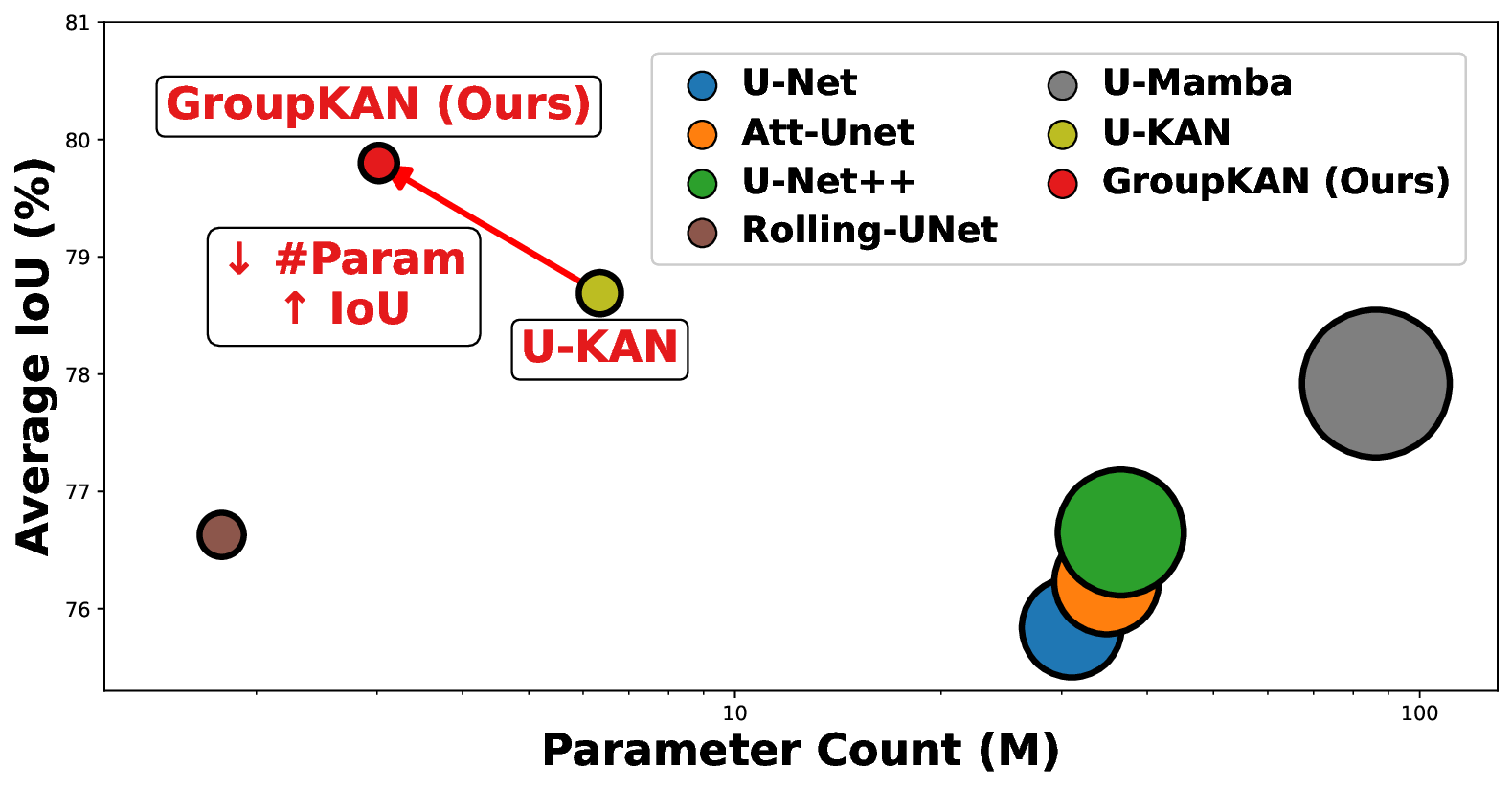} 
\caption{Accuracy-parameter efficiency trade-off. IoU (\%) vs. parameter count (M) across representative backbone models. Marker sizes are proportional to computational cost (GFLOPs).}
\label{GroupKAN_efficiency}
\end{minipage}\hfill
\begin{minipage}[c]{0.48\textwidth}
\centering
\makeatletter\def\@captype{table}\makeatother 
\footnotesize 
\caption{Impact of the number of groups ($g$) in GroupKAN. The configuration \( g = 16 \) is highlighted and achieves the best balance between segmentation performance and parameter efficiency.}
\label{tab:hyperparameters}
\begin{tabular}{c c c c}
\toprule
\textbf{Config.} & \textbf{IoU$\uparrow$} & \textbf{F1$\uparrow$} & \textbf{Params(M)}$\downarrow$ \\
\midrule
U-KAN & 64.15 & 77.10 & 6.35\\
$g=1$   & 63.69 & 77.46 & 6.76 \\
$g=4$   & 62.90 & 76.77 & 3.76 \\
$g=8$   & 65.68 & 78.95 & 3.27 \\
\rowcolor{gray!20}
$g=16$ & \textbf{67.66} & \textbf{80.52} & 3.02 \\
\bottomrule
\end{tabular}
\end{minipage}
\end{figure}

\noindent\textbf{Comparison with Representative Backbones.}
We evaluate GroupKAN against CNNs (U-Net~\cite{ronneberger2015u}, U-Net++~\cite{zhou2018unet++}), attention models (Att-Unet~\cite{oktay2018attention}, U-Mamba~\cite{ma2024u}), MLPs (U-NeXt~\cite{valanarasu2022unext}, Rolling-UNet~\cite{liu2024rolling}), and U-KAN~\cite{li2025u}. As shown in Tab.~\ref{tab:seg_comparison}, GroupKAN consistently outperforms all baselines. Qualitative comparisons (Fig.~\ref{fig:quality}) demonstrate that GroupKAN produces sharper boundaries with significantly fewer artifacts. Notably, it achieves a +1.11\% higher average IoU than U-KAN while halving both parameter count and FLOPs, a superior accuracy-parameter efficiency trade-off explicitly visualized in Fig.~\ref{GroupKAN_efficiency}. \\

\begin{figure}[htbp]
\centering
\includegraphics[width=.90\linewidth]{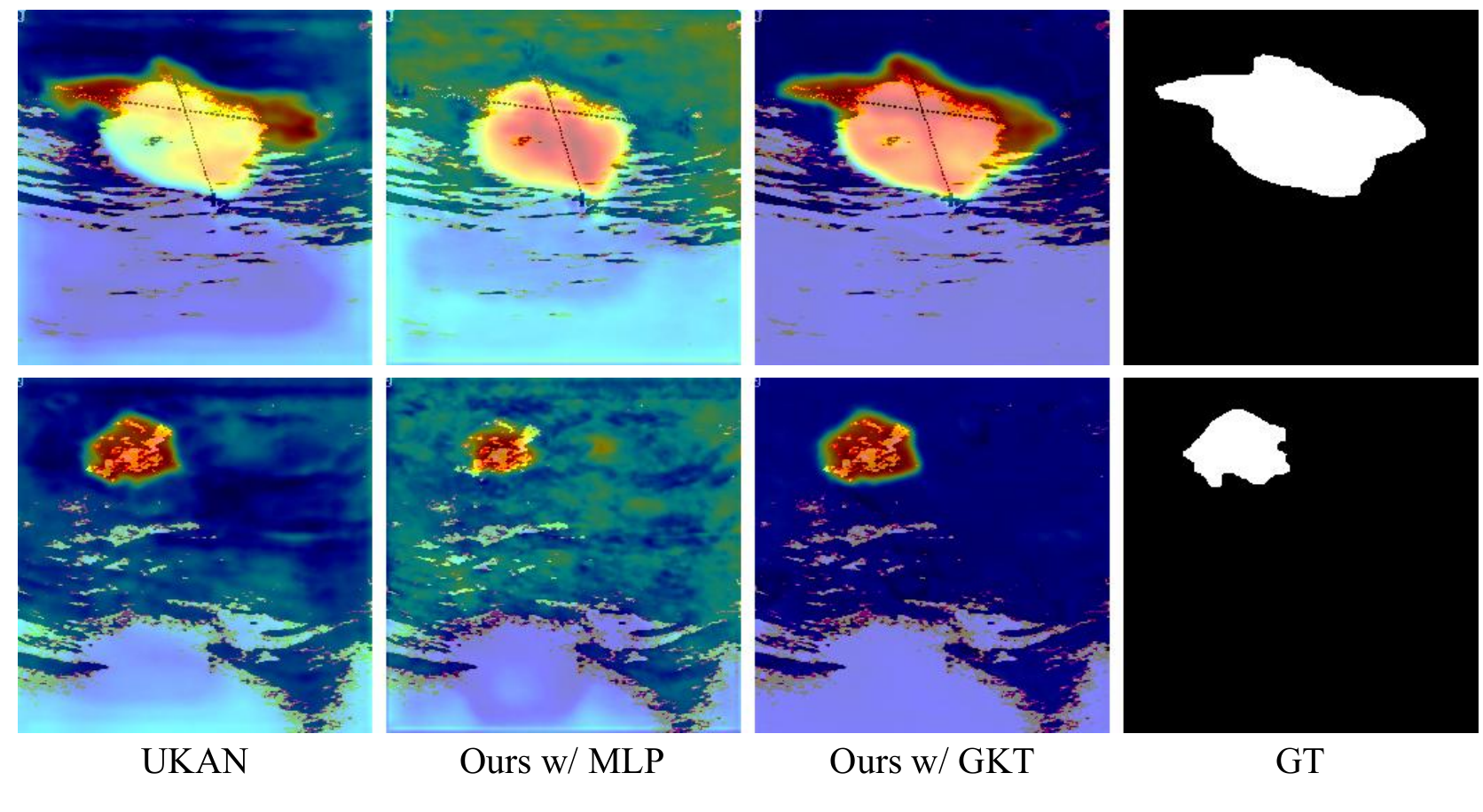} 
\caption{Channel activation maps for interpretability comparison. Note that ``Ours w/ GKT'' denotes the configuration equipped with our full proposed block, integrating both GKA and GKT. Compared to MLP and KAN baselines that struggle with precise localization, GKT produces sharply focused activations aligned with the ground truth.}
\label{interpretability}
\end{figure}

\noindent\textbf{Hyperparameter Configurations.} We evaluate the impact of the group number $g$ and compare our approach against the U-KAN (Table~\ref{tab:hyperparameters}). U-KAN and our $g=1$ variant demand the highest parameter counts (6.35M and 6.76M, respectively) yet yield suboptimal performance ($\sim$63--64\% IoU). As $g$ increases, the parameter count decreases monotonically. Ultimately, setting $g=16$ achieves the highest IoU (67.66\%) while compressing the parameter count to 3.02M. 
\begin{table*}[t]
\centering
\footnotesize 
\setlength{\tabcolsep}{3.5pt} 
\begin{minipage}[t]{0.46\textwidth}
\centering
\makeatletter\def\@captype{table}\makeatother 
\caption{Ablation on different feature modeling modules.}
\begin{tabular}{lcc}
\toprule
\textbf{Module} & \textbf{IoU}$\uparrow$ & \textbf{F1}$\uparrow$ \\ 
\midrule
None & 60.15 & 74.83 \\
KAN $\times 3$ & 66.08 & 79.39 \\
MLP $\times 3$ & 63.96 & 77.55 \\
\rowcolor{gray!20}
GKT $\times 3$ (Ours) & \textbf{67.66} & \textbf{80.52} \\ 
\bottomrule
\end{tabular}
\label{tab:effect_gkt}
\end{minipage}
\hfill
\begin{minipage}[t]{0.46\textwidth}
\centering
\caption{Comparison of different activation functions.}
\begin{tabular}{l|cc}
\toprule
\textbf{Activation} & \textbf{IoU}$\uparrow$ & \textbf{F1}$\uparrow$ \\
\midrule
None & 66.10 & 78.22 \\
ReLU \cite{nair2010rectified} & 66.63 & 79.24 \\
GoLU \cite{das2025golu} & 66.84 & 79.82 \\
\rowcolor{gray!20}
GKA (Ours) & \textbf{67.66} & \textbf{80.52} \\
\bottomrule
\end{tabular}
\label{tab:active}
\end{minipage}
\end{table*}

\subsection{Ablation Study}
\noindent\textbf{Effect of GKT.} We ablate GKT by removing it or replacing it with standard MLPs or unconstrained KANs. Quantitative results (Table~\ref{tab:effect_gkt}) reveal a clear hierarchy: GKT (67.66\% IoU) $>$ KAN (66.08\% IoU) $>$ MLP (63.96\% IoU) $>$ None (60.15\% IoU). Specifically, our proposed GKT yields a +1.58\% IoU and +1.13\% F1 improvement over the unconstrained KAN baseline. This confirms that the performance improvement comes fundamentally from the group-aware structural design rather than spline usage alone. The qualitative visual comparison in Fig.~\ref{interpretability} further demonstrates that GKT produces sharply focused activations better aligned with the corresponding ground truth.\\
\\
\noindent\textbf{Effect of GKA.} Compared to both fixed (ReLU~\cite{nair2010rectified}) and learnable (GoLU~\cite{das2025golu}) activations, our GKA consistently yields superior results, particularly in F1-score, which reflects better delineation of fine structures (see Table \ref{tab:active}).

\section{Conclusion}

We present GroupKAN, a lightweight architecture for medical image segmentation. To address the parameter explosion and overfitting risks of unconstrained KANs, we introduce group-aware spline modeling as a fundamental structural constraint. 
By integrating Grouped KAN Activation (GKA) and Transform (GKT), our framework efficiently models token-wise non-linearities and intra-group channel interactions, reducing practical model complexity by lowering parameter redundancy and improving efficiency. 
Across the BUSI, GlaS, and CVC benchmarks, GroupKAN consistently outperforms convolutional, attention-based, and standard KAN backbones, achieving superior accuracy while using less than half the parameters of U-KAN. 
Furthermore, qualitative analyses confirm that our design yields sharply localized activation maps, better aligned with the ground truth compared to MLP and KAN baselines.

\bibliographystyle{splncs04}
\bibliography{main}

\end{document}